\documentclass[fleqn,11pt]{wlscirep}
\usepackage[utf8]{inputenc}
\usepackage[T1]{fontenc}
\usepackage{bm}
\usepackage{float}
\usepackage{soul}
\graphicspath{{./pix/}}
\usepackage{subfig}
\usepackage{algorithm}
\usepackage{algorithmic}

\DeclareMathAlphabet{\mathcal}{OMS}{cmsy}{m}{n}

\title{Building Effective Large-Scale Traffic State Prediction System: Traffic4cast Challenge Solution}
\author[1,a,*]{Yang Liu}
\author[2,b,*]{Fanyou Wu}
\author[3]{Baosheng Yu}
\author[1]{Zhiyuan Liu}
\author[4]{Jieping Ye}
\affil[1]{Southeast Univeristy, School of Transportation, Nanjing, China}
\affil[2]{Purdue University, Department of Forestry and Natural Resource, West Lafayette, USA }
\affil[3]{The University of Sydney, UBTECH Sydney AI Centre, School of Computer Science, Sydney, Australia }
\affil[4]{DiDi Chuxing, Didi AI Labs, Beijing, China}
\affil[a]{Email: 230179629@seu.edu.cn}
\affil[b]{Email: wu1297@purdue.edu}
\affil[*]{Equal Contribution}

\begin{abstract}
How to build an effective large-scale traffic state prediction system is a challenging but highly valuable problem. This study focuses on the construction of an effective solution designed for spatio-temporal data to predict large- scale traffic state. Considering the large data size in Traffic4cast Challenge and our limited computational resources, we emphasize model design to achieve a relatively high prediction performance within acceptable running time. We adopt a structure similar to U-net and use a mask instead of spatial attention to address the data sparsity. Then, combined with the experience of time series prediction problem, we design a number of features, which are input into the model as different channels. Region cropping is used to decrease the difference between the size of the receptive field and the study area, and the models can be specially optimized for each sub-region. The fusion of interdisciplinary knowledge and experience is an emerging demand in classical traffic research. Several interdisciplinary studies we have been studying are also discussed in the Complementary Challenges. The source codes are available in \url{https://github.com/wufanyou/traffic4cast-TLab}.
\end{abstract}
\begin{document}

\flushbottom
\maketitle
\thispagestyle{empty}
\section{Introduction}

Real-time traffic state prediction is an essential component for traffic control and management in an urban road network. The ability to predict future traffic state (e.g., flow, speed) can help improve traffic conditions, fleet organization, utilization rate, and social welfare~\cite{ke2017short,yao2018deep}.
Essentially, the traffic prediction is a time series problem, which is performed based on the changes of historical demand. A representative time-series prediction tool is the recurrent neural network (RNN), along with its diverse variants~\cite{liu2019deeppf,fu2016using}. Apart from the temporal dimension, the correlation in the spatial dimension is also extensively incorporated by many works. Regions that are close to each other or share similar land-use structures may exhibit a homogeneous demand pattern. Techniques widely applied in computer vision like convolutional neural network (CNN)~\cite{yao2018deep,zhang2017deep} and the emerging graph-based networks~\cite{geng2019spatiotemporal,li2017diffusion,Pan:2019:UTP:3292500.3330884,yu2018spatio} are often adopted. Furthermore, multi-source data are also introduced in some literature to allow for the external influencing factors, such as weather conditions and neighboring points-of-interest~\cite{koesdwiady2016improving,liao2018deep}.

\section{Data Description and Problem Definition}

\begin{figure}[bt]
\centering
\subfloat[{\label{traffic4cast:Berlin}Berlin}]{
  \includegraphics[width=0.3\linewidth]{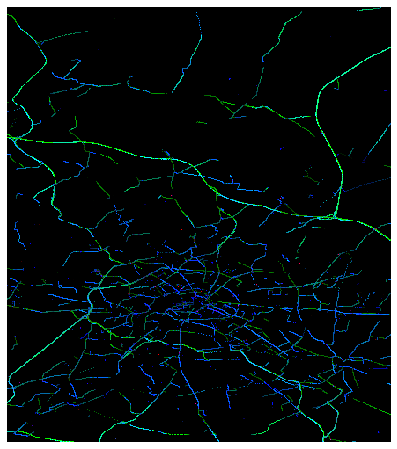}
}
\subfloat[{\label{traffic4cast:Istanbul}Istanbul}]{
  \includegraphics[width=0.3\linewidth]{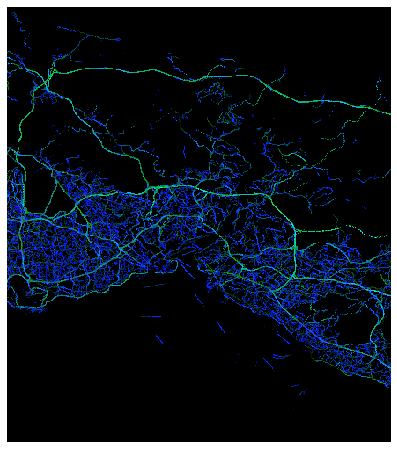}
}
\subfloat[{\label{traffic4cast:Moscow}Moscow}]{
  \includegraphics[width=0.3\linewidth]{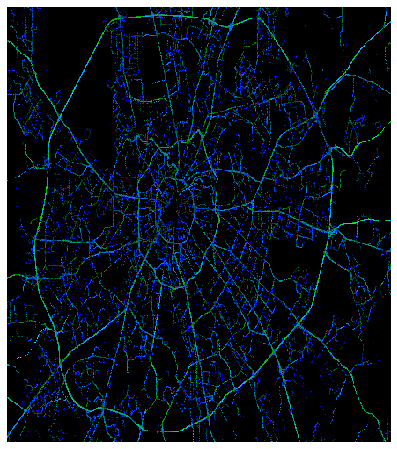}
}\\
\caption{\label{traffic4cast:samples} Sample Images from traffic4cast data set for May 1\textsuperscript{st}, 2018. In general, the greener the color, the heavier the road.}
\end{figure}

The organizer provides industrial-scale real-world data for 3 full cities (\hyperref[traffic4cast:Berlin]{Berlin}, \hyperref[traffic4cast:Istanbul]{Istanbul} and \hyperref[traffic4cast:Moscow]{Moscow}) throughout a year. The organizer divides the whole city into a $436\times495$ grid, each pixel of which represents a region of $100\,m \times 100\,m$. There are a totally of 288 frames each day, and each frame of the raw data represents 5-minutes aggregated information with three channels, \textit{i.e.}, traffic volume, speed, and direction, respectively. The sample images are illustrated in \hyperref[traffic4cast:samples]{Figure}~\ref{traffic4cast:samples}. The data of volume and speed are scaled to $[0,255]$ through a min-max scaler. Regarding the direction data, they are obtained by binning probes into four heading directions of North-East $(0, 90)$, South-East $(90, 180)$, South-West $(180, 270)$ and North-West $(270, 359]$. In this way, the raw heading values $[0, 359]$ are mapped as four distinct values of $\{1, 85, 170, 255\}$ in the order of $(NW, NE, SW, SE)$, and the major heading bin, \textit{e.g.} the bin containing the largest number of points, is selected. Here, missing values are also represented by 0.
\vspace{1ex}

\textbf{Problem:} This challenge can be split into three sub-tasks, i.e., use the given data to predict speed, flow, and direction at each pixel separately in the given time intervals, which is a typical time series prediction problem. Pixel-wise mean squared error (MSE) is used to evaluate the performance for ranking the submitted prediction results.

\section{Solution}

\subsection{Network}

\textbf{Increasing the receptive field.} One important attribute of the data is the spatial correlation. This means that the traffic state in a specific region can be affected by the adjacent regions. For example, when traffic congestion occurs in a link of the urban road network, it might propagate upstream, thus causing traffic congestion in neighboring areas. To capture this phenomenon, a large receptive field is theoretically better. A simplest way to increase receptive field is to increase kernel size of convolutional operation. However, due to the limit of computing devices, increasing the receptive field through enlarging the convolutional kernels is not applicable. Therefore, pooling operation is a more feasible way to increase the receptive field. Considering that this task is similar to pixel-wise image segmentation, we adopt a structure similar to U-net~\cite{ronneberger2015u}. Moreover, the inputs are cropped to further decrease the difference between the size of the receptive field and the study area (see \hyperref[traffic4cast:crop]{section}~\ref{traffic4cast:crop} for more details).

\textbf{Use mask instead of spatial attention.} To address the data sparsity, the introduction of spatial attention might be possible, but it will lead to a sharp increase in training time. Therefore, a \textit{mask} is designed to improve efficiency. We concatenate one more channel to the last layer before the $1\times1$ \textsc{Conv}, which checks whether the mean of input is equal to zero or not.

\subsection{Feature Engineering}

\textbf{Convert features into channels.} Essentially, this task is a time-series prediction problem. For one-dimensional time-series prediction, a typical solution is to transform the problem into a supervised learning problem with the help of feature engineering. Referring to this idea, the features are input into the model as different channels. Limited by the computing devices (\textit{i.e.}, $4\times2080\text{Ti}$ GPUs), there are 101 input features (\textit{i.e.}, 101 channels). More features can usually bring higher accuracy. 

\textbf{Time smoothness features.} Time series data are intrinsically continuous rather than discrete. Hence, time series seldom produce sharp variations, which means that the traffic state at a certain time interval has a close resemblance to traffic state at the adjacent time interval. The feature crosses are also considered. Therefore, in each prediction subtask, the data of all three subtasks from the previous 12 time intervals are used. A total of $12\times3$ channels are used from  $[T-12,T-1]$ with the same day, where $T$ denotes the predicted time index, and the number 3 indicates the data of three subtasks.

\textbf{Periodic features.} The similarity between the traffic states of two different days can be attributed to the periodic characteristics of traffic states, which typically repeat every 24 hours. We use in total $16\times3$ channels from $\{D-14\} \bigcup[D-7, D-1]\bigcup [D+1, D+7]\bigcup\{D+14\}$ with the same time index, where $D$ is the predicted day. To remove random noises, moving windows with the windows size of 13 are applied to smooth each time index.

\textbf{Statistical features.} We use in total $5\times3$ channels from 5 average given period on the same day.

\textbf{Time-dependent features.} Two channels in total are used to represent month and week average in the predicted time index and corresponding task.

\subsection{\label{traffic4cast:crop} Region Cropping}

\begin{figure}[ht]
  \centering
  \includegraphics[width=0.3\linewidth,angle=-90]{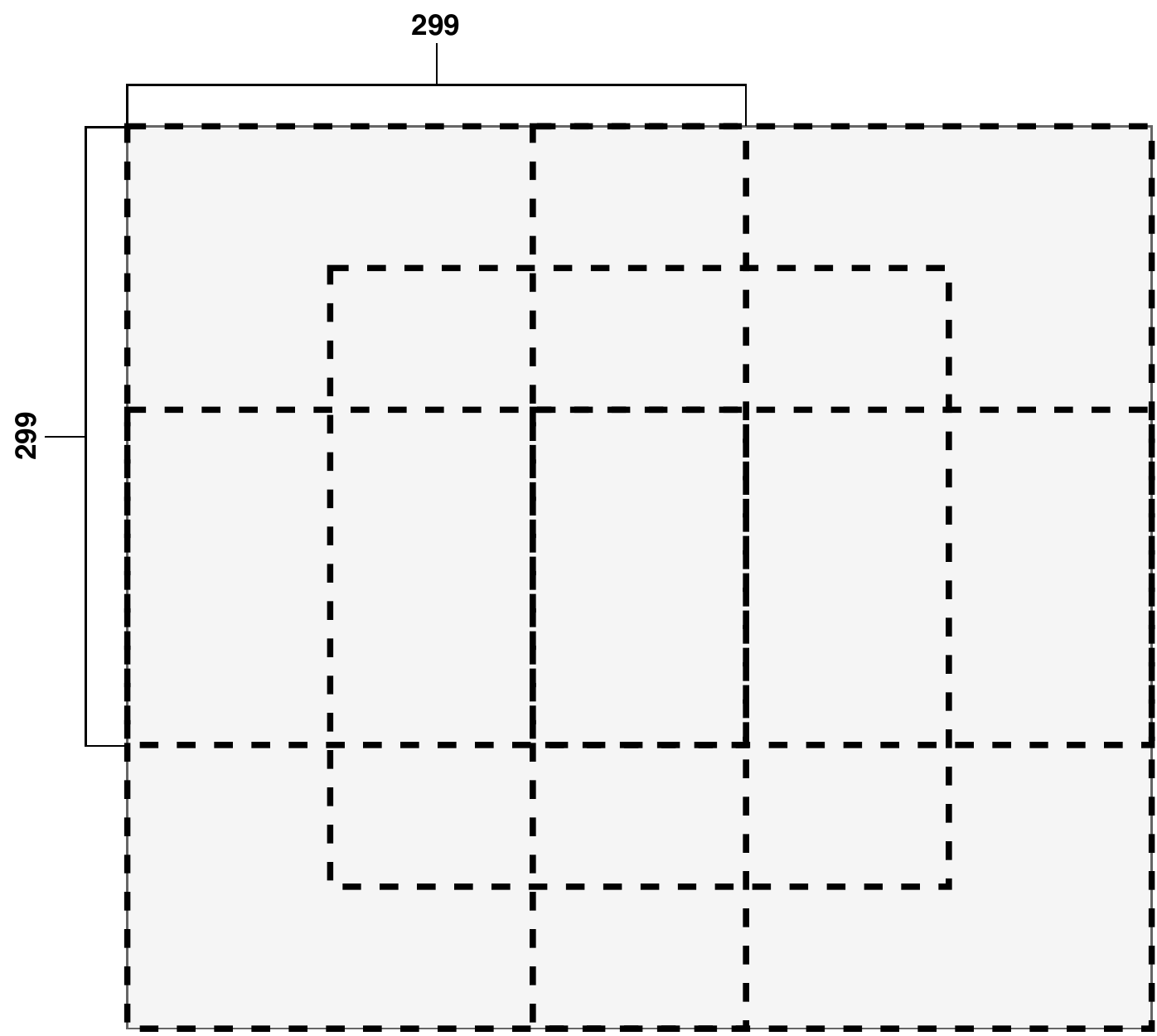}
  \caption{ \label{fig:crop} Schematic diagram of region cropping}
\end{figure}

We further divide the study area into five sub-regions with size $299\times299$ (see \hyperref[fig:crop]{Figure}~\ref{fig:crop}). Region cropping allows our model to be trained on weak computing devices. As mentioned above, it decreases the difference between the size of the receptive field and the study area. In addition, the models can be specially optimized for each sub-region, as the optimal parameters for the whole city may not be the best parameters suited for each sub-region. We spliced the prediction results of each sub-region into a complete region, and the overlapping portions were replaced with average values. To sum up, in our final solution, we trained in total $3^\text{cities}\times 3^\text{subtasks} \times 5^\text{subregions} = 45$ models. Here, it should be pointed out that our models can be fine-tuned and actually efficient in training.

\subsection{Training}
We train all models in following procedure. First, for each city, we train a model from scratch for direction prediction task of a $299\times299$ subregion. Then using those parameters as pre-trained models, we fine-tune all the rest models. The fine-tune step starts from a larger learning rate, more data, and no data augmentation to lower learning rate, fewer data, and more data augmentation. The training process is summarized in 
\hyperref[alg:Training]{Algorithm}~\ref{alg:Training} as below.

\begin{algorithm}[ht]
\caption{\label{alg:Training}Training}
\begin{algorithmic}[1] 

\FOR{CITY \textbf{in} \{Berlin, Istanbul, Moscow\} }
\STATE Train a model from scratch for direction prediction task with SUB\_REGION$=0$.
\FOR{SUB\_REGION \textbf{in} $[0,4]$ }
\STATE Use the pre-trained model above, fine-tune the models for the other tasks.
\ENDFOR
\FOR{SUB\_REGION \textbf{in} $[0,4]$ }
\STATE Fine-tune all the models with lower learning rate, fewer data, and data augmentation.
\ENDFOR
\ENDFOR
\STATE \textbf{return} $\bm x$
\end{algorithmic}
\end{algorithm}


\section{Complementary Challenges}
\subsection{Alternative Metrics}
Pixel-wise mean squared error (MSE) is used to evaluate the performance for the ranking of submitted prediction results. The MSE is not the idea loss for this data set due to the sparsity of data and the improper coding of the heading channel. For Moscow, the MSE of the heading channel is approximate $3.1\times10^{-2}$ while speeding, and volume channel are $1 \times 10^{-4}$ and $5 \times 10^{-3}$ respectively. This magnitude difference also happened in Berlin and Istanbul. It is clear that the heading channel dominates overall scores comparing to the leader-board score difference in $10^{-4}$. In other words, a model which predicts better heading channel will rank higher in the leader board. If we direct train the model with MSE loss, though the performance is the best regarding lead board ranking, the predicted values are meaningless. \hyperref[traffic4cast:predvstreu]{Table}~\ref{traffic4cast:predvstreu} is an example: our best model tries to predict large value more conservative (trend to out-put small values and made frame unrealistic and blurry ) because of the punishment of square error. This characteristic hinders this model to real-world applications when detailed general directions $(NW, NE, SW, SE)$ are more critical. In real-world traffic prediction systems, MSE is not also welcome because it cannot give a sense of how accurate prediction is. Mean Absolute Error (MAE) and Mean Absolute Percentage Error (MAPE) is more often used in traffic prediction tasks. Given the above explanations, we consider to apply a multi-loss combine cross-entropy and weight MAPE together, and define the new loss function $\mathcal{L}$ as:

 \begin{equation}
     \mathcal{L}=w_{\text{heading}} \cdot \text{CE}(\text{heading})+w_{\text{volume}} \cdot \text{MAPE}(\text{volume})+w_{\text{speeding}} \cdot \text{MAPE}(\text{speeding})
 \end{equation} Here $\text{CE}(\cdot)$ represents the 2D cross-entropy loss with 5 classes $(0, NW, NE, SW, SE)$ while $\text{MAPE}(\cdot)$ is the 2D Mean Absolute Percentage Error.
 

\begin{table}[b]
\caption{ \label{traffic4cast:predvstreu} Average Predicted Values ($\bar{I}_{\text{predict}}$) verse its true value ($I_{\text{true}}$) for heading channel. We select (Moscow, 20181106, 258-260) as an example.}
  \centering
  \begin{tabular}{ll}
    \toprule
    $I_{\text{true}}$  & $\bar{I}_{\text{predict}}$ \\
    \midrule
    0&8.08\\
    1&58.63\\
    85&73.44\\
    170&93.17\\
    255&117.73\\
    \bottomrule
  \end{tabular}
\end{table}

\subsection{Multi-source Data Fusion and Online Learning}

Multi-source data fusion is fashion and often be reported as useful in the research fields, but we did not embed any external information to the models for the final solution. We report that embedding external information \textit{e.g.}, weather, holiday, and the day of week might lead to over-fitting. We tried to embed 20 external features into 64 dimension vector and add them to the output of the first $3\times3$ \textsc{Conv} of U-net, but the MSE error increased significantly when other features remain unchanged. 

Here we propose a method that could apply external information to our model. For each predicted day, the model will be fine-tuned one more epoch using similar data of the predicted day. The similarities between data are measured by time, weather, and holiday features. If we just use the time to rank the similarity, then this method downgrade to an online learning method - the model tries to optimize recent inputs instead of global ones. By using this method, we had a deduction of $10^{-4}$ to loss. 

\subsection{Spatio-Temporal Ensemble Method}

The motivation of spatio-temporal ensemble method stems from the large-scale spatio-temporal predictions performed in many online taxi-hailing service providers, \textit{e.g.,} Uber, Didi, and Lyft. These predictions are usually completed by different teams from different domains, with different models. However, every single model may have some inherent defects due to insufficient information, and hence it is rational to ensemble the results of these models to improve the overall performance. It remains a challenging but essential issue to make an effective combination of multiple base models while leveraging the spatio-temporal information. To the best of our knowledge, the research in this vein is very sparse.

The prediction outcome can help the online car-hailing platforms accurately dispatch the fleet and lower the vehicle vacancy ratio, thus serving more passengers with their trip requests. Notably, even if the average prediction error reduction is small, compared with the optimal base model, the cumulative number of orders for a long period, say, one year, will be considerable in that the whole city can be seen as thousands of pixels, and DiDi and Uber run their business in hundreds of cities. Therefore, we design an attention-based deep ensemble net in dealing with the problems mentioned above~\cite{liu2019Attention}.

\subsection{Personalized Intelligent Transportation Prediction Systems}

In our recent study, we managed to introduce the concept of personalization in the recommendation system to the prediction problem in intelligent transportation systems, and a more intelligent urban transportation system, which conforms to the individual preferences and is appropriate for regional traffic conditions, can thus be designed. Therefore, we explore personalization in the application of intelligent transportation systems. Aiming at the two common personalized demands in intelligent transportation systems, we define the two types of personalization in intelligent transportation systems on the basis of different application scenarios: User Personalization and Geographic Personalization~\cite{liu2019Exploring}. Interdisciplinary knowledge and experience fusion is urgently needed classical traffic research. In addition to the construction of personalized intelligent transportation systems, there is also great potential for adversarial examples and transfer learning. 

\clearpage
\bibliography{main}
\end{document}